\documentclass[10pt,twocolumn,letterpaper]{article}

\usepackage{iccv}
\usepackage{times}
\usepackage{epsfig}
\usepackage{graphicx}
\usepackage{amsmath}
\usepackage{amssymb}
\usepackage{graphicx}
\usepackage{diagbox}
\usepackage{rotating}
\usepackage{multirow}
\usepackage[normalem]{ulem}
\usepackage{caption}
\usepackage{subcaption}
\usepackage{booktabs}
\usepackage{colortbl}
\usepackage{array}
\usepackage{xcolor}
\usepackage{times}  
\usepackage{helvet}  
\usepackage{courier}  
\usepackage[hyphens]{url}  
\usepackage{graphicx} 
\usepackage{booktabs}
\usepackage{tabularray}
\usepackage{xcolor}
\usepackage{caption} 
\usepackage{algorithm}
\usepackage{algorithmic}
\usepackage{amsmath}
\usepackage{amssymb}
\usepackage{amsfonts}
\usepackage{newfloat}
\usepackage{listings}
\usepackage[accsupp]{axessibility}  

\usepackage[pagebackref=true,breaklinks=true,letterpaper=true,colorlinks,bookmarks=false]{hyperref}

\iccvfinalcopy 

\def\iccvPaperID{5954} 

\ificcvfinal\fi

\newcommand\blfootnote[1]{%
  \begingroup
  \renewcommand\thefootnote{}\footnote{#1}%
  \addtocounter{footnote}{-1}%
  \endgroup
}

\begin{document}
\title{TC-LLaVA: Rethinking the Transfer from Image to Video Understanding with Temporal Considerations}

\author{Mingze Gao$^{1,2,3,\dag}$\quad  Jingyu Liu$^{2}$ \quad  Mingda Li$^{2}$\quad  Jiangtao Xie$^{4}$ \quad Qingbin Liu$^{2}$ \\ \quad Bo Zhao$^{2}$ \quad  Xi Chen$^{2}$ \quad  Hui Xiong$^{1,3, *}$ \\
$^{1}$ The Hong Kong University of Science and Technology (Guangzhou), China\\
\quad \qquad $^{2}$Tencent PCG
\\ \qquad $^{3}$The Hong Kong University of Science and Technology, China\\
\qquad $^{4}$ Dalian University of Technology, China
}

\maketitle
\begin{abstract}
\blfootnote {$^\dag$ This work was done when Mingze Gao was an intern at Tencent PCG. $^*$ Corresponding authors}
Multimodal Large Language Models (MLLMs) have significantly improved performance across various image-language applications. Recently, there has been a growing interest in adapting image pre-trained MLLMs for video-related tasks. However, most efforts concentrate on enhancing the vision encoder and projector components, while the core part, Large Language Models (LLMs), remains comparatively under-explored. In this paper, we propose two strategies to enhance the model's capability in video understanding tasks by improving inter-layer attention computation in LLMs. Specifically, the first approach focuses on the enhancement of Rotary Position Embedding (RoPE) with Temporal-Aware Dual RoPE, which introduces temporal position information to strengthen the MLLM's temporal modeling capabilities while preserving the relative position relationships of both visual and text tokens. The second approach involves enhancing the Attention Mask with the Frame-wise Block Causal Attention Mask, a simple yet effective method that broadens visual token interactions within and across video frames while maintaining the causal inference mechanism. Based on these proposed methods, we adapt LLaVA for video understanding tasks, naming it Temporal-Considered LLaVA (TC-LLaVA). Our TC-LLaVA achieves new state-of-the-art performance across various video understanding benchmarks with only supervised fine-tuning (SFT) on video-related datasets.

\end{abstract}

%

\section{Introduction}
By leveraging vast open-source and AI-generated datasets~\cite{lin2014microsoft, schuhmann2022laion, chen2023sharegpt4v}, along with the impressive development of large language models such as GPT~\cite{achiam2023gpt}, LLaMA~\cite{touvron2023llama}, and GLM~\cite{du2021glm}, Multimodal Large Language Models (MLLMs) have demonstrated remarkable proficiency in image comprehension tasks~\cite{liu2024visual, liu2024improved, li2023blip, zhu2023minigpt}. Given the powerful capabilities of image-pretrained MLLMs, a recently emerging research focus is on transferring these models from single-image tasks to video understanding.

\begin{figure}[ht]
\centering
   \includegraphics[width=1.0\linewidth,trim=0 20 0 0,clip]{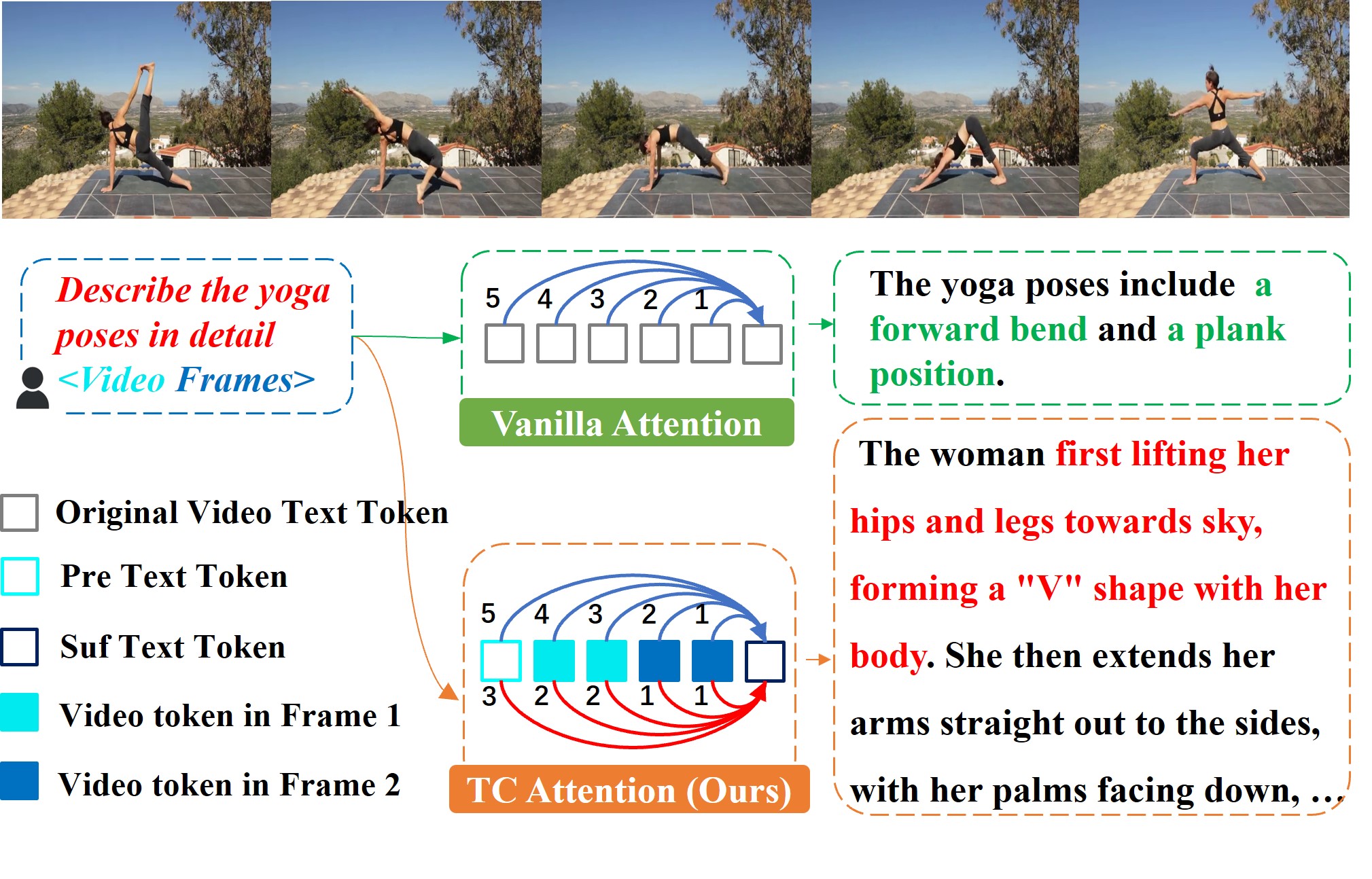}
   \caption{Video language processing with LLaMA~\cite{touvron2023llama} and our TC-LLaVA, where arrows represent the attention interactions with this token, and numbers indicate the relative positional distance between tokens. Vanilla Attention uniformly encodes and applies attention interactions to both visual and text tokens. The proposed TC Attention incorporates temporal information encoding and differentiates interactions between visual tokens within and across frames, which are indicated by different colors.}
\label{Framework 1}
\end{figure}

Recently, various approaches~\cite{maaz2023video, yang2022zero, zhang2024llavanextvideo} have tended to treat a video as a series of concatenated frames in the spatial dimension, thereby transferring video-related tasks back to image-related tasks. However, these methods face two issues as they treat text and visual tokens as the same modality and fed them into the LLMs as a unified input. Firstly, utilizing LLMs' vanilla attention mechanism to uniformly process all tokens overlooks the distinct interactions between visual tokens within individual video frames and those across different frames. Secondly, it neglects the temporal information inherent in the video input, which is crucial for video understanding tasks. Consequently, the constructed video MLLM fails to effectively summarize the dynamic events occurring within videos, reducing the analysis to single frames as if they were still images. For instance, it fails to adequately capture and detail the complex motion changes of the primary subject in the video, particularly in activities such as dancing or gymnastics. This deficiency ultimately results in inaccurate or 'hallucinatory' responses by the model, as depicted in Figure~\ref{Framework 1}.

In this paper, we propose Temporal-Considered (TC) LLaVA, a novel video-language framework designed to address the aforementioned issues. The primary innovation is to enhance the temporal awareness of MLLMs and distinguish the attention interactions between text and video modalities through two core strategies. First, we introduce Temporal-Aware Dual RoPE, which assigns each token an independent position id with the original RoPE to preserve global relative positional relationships, while incorporating  temporal-aware RoPE to assign the same position id to visual tokens within the same frame and to encode inter-frame relationships to capture the temporal dynamics of videos, as shown in Figure~\ref{Framework 1}. Additionally, we design three different attention masks to optimize token interaction strategies in attention computation, accounting for the distinct characteristics of visual and text tokens. Finally, we select the Frame-wise Block Causal Attention Mask to replace the original causal attention mask, enhancing interaction between visual tokens within and across frames while preserving the causal reasoning paradigm, making it more suitable for causal language model inference.

To verify the effectiveness of our TC-LLaVA, we evaluate the model on extensive video benchmarks, including MSVD~\cite{xu2016msr}, MSRVTT~\cite{xu2016msr}, ActivityNet~\cite{caba2015activitynet}, TGIF~\cite{li2016tgif}, VCGbench~\cite{maaz2023video} and MVbench~\cite{li2023mvbench}. Comparing with the latest video MLLMs, TC-LLaVA achieves new state-of-the-art performance on these benchmarks at the same model scales, demonstrating the benefits of enhancing visual token interactions within and across frames, as well as the importance of incorporating temporal information in video analysis.

\begin{figure*}[ht]
\centering
   \includegraphics[width=1.0\linewidth,trim=0 0 0 0,clip]{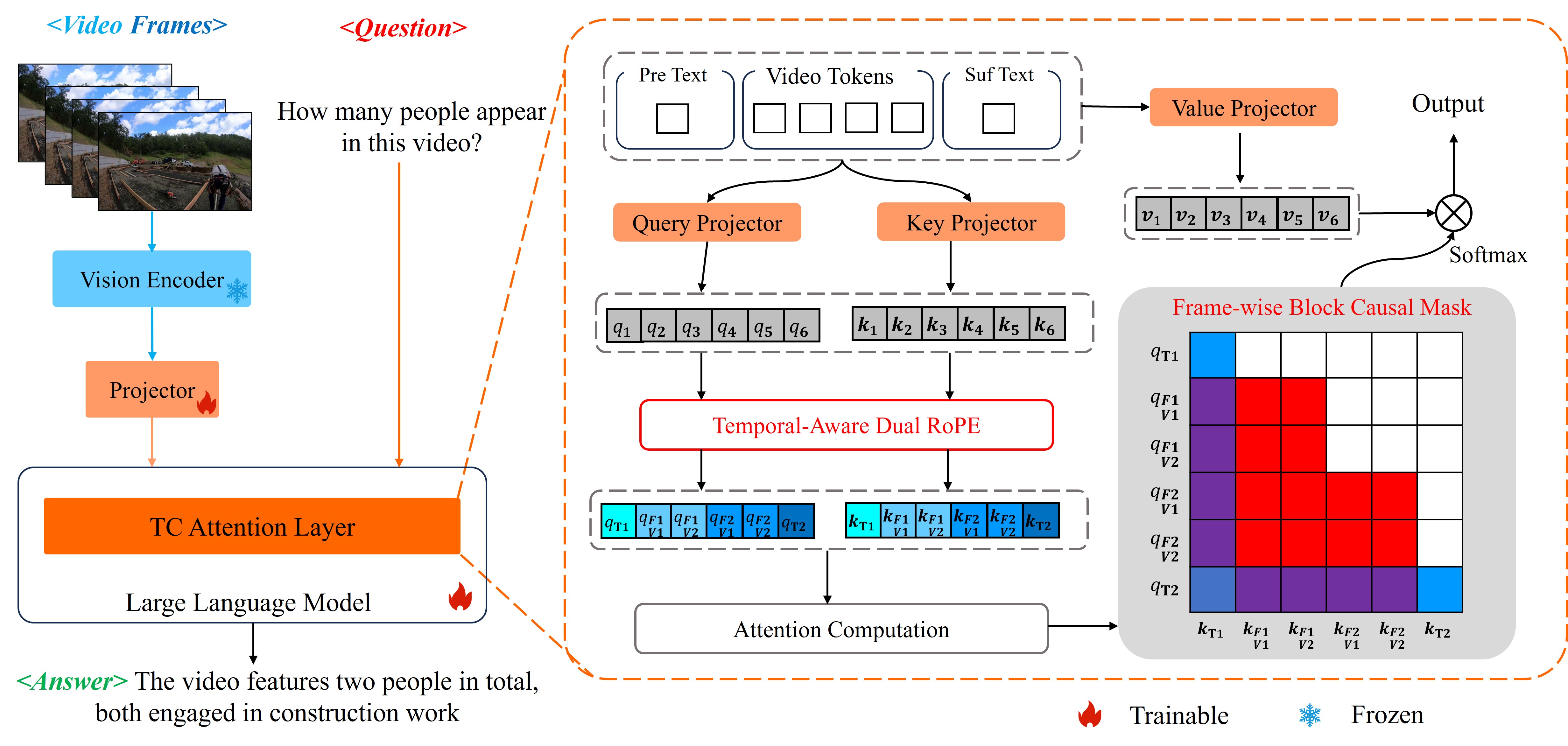}
   \caption{\textbf{The framework of TC-LLaVA}. During SFT stage, the projector and the language model (LLM) are unfrozen, while the visual encoder remains frozen. The right part illustrates our TC-attention mechanism in each Transformer layer. After applying Temporal-Aware Dual RoPE, both visual and text tokens acquire additional temporal positional information while preserving the global relative positional relationships. Frame-wise Block Causal Mask aims to enhance the visual tokens interactions within and across frames.}
\label{Framework 2}
\end{figure*}

\section{Related Work}
\subsection{Attention in Vision and Language Models}
The introduction and evolution of the attention mechanism have significantly enhanced model performance in natural language processing (NLP) and computer vision (CV). The earliest attention mechanism by ~\cite{bahdanau2014neural} allowed machine translation models to assign different weights to input sentence parts, improving translation accuracy. ~\cite{vaswani2017attention} introduced the Transformer model, which uses a self-attention mechanism to enable parallel processing and superior long-range dependency modeling, achieving significant results in multiple NLP tasks. To further optimize the attention computation, \cite{shaw2018self} propose Relative Position Encoding (RPE) to improve the token interaction by introducing extra position information. Recently, Rotary Position Embedding (RoPE)~\cite{su2024roformer} is designed for the interaction limitation of RPE by leveraging complex number rotations. In CV, attention mechanisms have proven effective with models like Non-local Neural Networks by~\cite{wang2018non} and Vision Transformer (ViT)~\cite{dosovitskiy2020image}, which first applies the Transformer architecture to image classification tasks. There have also been numerous advancements~\cite{liu2021swin, xie2021segformer, liu2022swin} in attention mechanisms that continually improve the performance of Transformer-based models, enhancing their ability to capture essential features and increasing computational efficiency. Our work continues to delve deeply into improving attention computation in the multimodal domain of video and text, and we propose the TC-Attention method to achieve this goal.

\subsection{Video Multimodal Large Language Models}
Video Multimodal Large Language Models (Video MLLMs) operate by aligning modalities and performing instruction fine-tuning on video data, enabling them to generate responses based on user instructions and input video streams. Recently, Video MLLMs have experienced rapid development. One significant milestone in this field is BLIP2~\cite{li2023blip}, which integrates a frozen vision encoder with a Q-Former to enhance video processing efficiency, demonstrating remarkable zero-shot capabilities in Video Question Answering (VQA) and outperforming existing techniques. Video-ChatGPT~\cite{maaz2023video} introduced video instruction tuning and created a high-quality instructional dataset, setting a new standard for video-based text generation benchmarks. VideoChat~\cite{2023videochat} employed cross-attention mechanisms to condense visual tokens and align user queries with the dialogue context, enhancing interpretative capabilities. Building on this, VideoChat2~\cite{li2023mvbench} refined the approach with a multi-stage bootstrapping technique focused on modality alignment and instruction tuning, utilizing a robust collection of high-quality video data. Chat-UniVi~\cite{jin2024chat}  processes longer videos by introducing a method for compressing tokens in both the spatial and temporal dimensions. LLaMA-VID~\cite{li2023llama} introduced an innovative dual-token approach that effectively condenses video representations by segregating context and content tokens, allowing for more efficient compression. VideoLLaMA and VideoLLaMA2~\cite{zhang2023video, cheng2024videollama} enhances video understanding by incorporating audio modality information and utilizing a Spatial-Temporal Convolution (STC) connector. ST-LLM~\cite{liu2024st} intruduce a dynamic masking strategy into MLLM. PLLaVA~\cite{xu2024pllava} explore the Image-pretrained LLaVA into video tasks with simple spatial pooling. In this paper, we introduce TC-LLaVA, which considers the differences in visual token interactions within and across frames, and directly incorporates temporal position into the causal attention computation to enhance the understanding of model.

\section{Method}
\subsection{Preliminary: Introducing Position Embeddings}
While Relative Position Encoding (RPE)~\cite{shaw2018self} incorporates relative positional information into the attention mechanism through a position bias element-addition computation with inter-layer attention map, this approach may limit interaction with attention weights and, consequently, hinder the effective utilization of relative positions. To address this limitation, RoFormer~\cite{su2024roformer} introduces RoPE, a novel method that more effectively incorporates relative positional information by leveraging complex number rotations.

\noindent Specifically, when computing the attention map, the RoPE (Rotary Positional Encoding) technique introduces the multiplication of Euler's formula \( e^{i\theta} \) to the query and key vectors as a relative position embedding. For instance, when considering the \(n\)-th and \(m\)-th query and key vectors \( q_n \) and \( k_m \) in \( \mathbb{R}^{1 \times d_{head}} \), RoPE is applied as follows:
\begin{align}\label{eq:1}
\mathbf{q}'_n = \mathbf{q}_n e^{i n \theta}, \quad \mathbf{k}'_m = \mathbf{k}_m e^{i m \theta}.
\end{align}

\noindent Then, the $(n, m)$-th component of the attention matrix is calculated as:
\begin{align}\label{eq:2}
A_{(n,m)} = \text{Re}[\mathbf{q}'_n \mathbf{k}'_m{}^*] = \text{Re}[\mathbf{q}_n \mathbf{k}_m^* e^{i (n-m) \theta}],
\end{align}
where $\text{Re}[\cdot]$ denotes the real part of a complex number and $^*$ denotes the complex conjugate. By multiplying complex rotations $e^{i \theta n}, e^{i \theta m}$ depending on token position $(n, m)$, RoPE injects relative positions $(n - m)$ into the attention matrix in a rotational form. In practical implementation, RoPE~\cite{su2024roformer} converts the vectors \( q_n \) and \( k_m \)  from $\mathbb{R}^{1 \times d_{\text{head}}}$ to complex vectors $\bar{q}_n$ and $\bar{k}_m$ in $\mathbb{C}^{1 \times (d_{\text{head}} / 2)}$. This is achieved by treating the $(2t)$-th dimension as the real part and the $(2t + 1)$-th dimension as the imaginary part, where $t \in {0, 1, \ldots, d_{\text{head}} / 2}$. This method results in the same attention values as $\mathbf{q}_n \mathbf{k}_m^T = \text{Re}[\bar{\mathbf{q}}_n \bar{\mathbf{k}}_m^*]$ while reducing computational overhead. Additionally, RoPE employs multiple frequencies $\theta_t$ through the channel dimensions of the query and key vectors as follows:
\begin{align}\label{eq:3}
\theta_t = 10000^{-t / (d_{\text{head}} / 2)},
\end{align}

\noindent This approach allows for more effective integration of relative positional information within the attention mechanism, enhancing the model's capability to process and understand sequential data.

\subsection{Temporal-Aware Dual RoPE}
In the RoPE used by most current video-language large language models, the relative distance between the \(m\)-th text token \(T_m\) and the \(z\)-th visual token in the \(n\)-th frame \(F_n V_z\) is defined as Eqn~\ref{eq:4}. Each text and visual token is treated as an independent position and assigned a unique position id for embedding. However, this position embedding method fails to distinguish visual tokens within and across different video frames, thereby neglecting the crucial temporal information necessary for effective video understanding tasks. Furthermore, as visual tokens constitute a significant proportion of the total tokens in video understanding tasks, the relative distance \(P(T_m) - P(F_n V_z)\) between the generated text tokens and the visual tokens may become substantial. This increased distance can impair the model's ability to fully comprehend the visual information, leading to "hallucinated" responses~\cite{ma2023vista}. 
\begin{align}\label{eq:4}
A_{(q_{T_m}, k_{F_n V_z})} = \text{Re}[\mathbf{q}_{T_m} \mathbf{k}_{F_n V_z} e^{i (P(T_m) - P(F_n V_z)) \theta}],
\end{align}

To address this limitation, we propose a Temporal-Aware Dual Rotary Positional Embedding (TAD-PoPE). It includes one RoPE that retains the global relative position relationships of the visual and textual tokens, and an additional time-aware RoPE to incorporate temporal information pertinent to the video frames. Specifically, in contrast to the original position ids, the additional RoPE ensures that visual tokens within the same video frame share the same position id. Meanwhile, the temporal order is maintained across different frames, with the position ids incrementing accordingly. The proposed temporal position id is defined as follows:

\begin{align}\label{eq:5}
\mathbf{I}_{\text{t}}(n) = 
\begin{cases}
n, & \text{if } n < v_{\text{s}}, \\
v_{\text{s}} + \left\lfloor \frac{n - v_{\text{s}}}{m} \right\rfloor, & \text{if } v_{\text{s}} \leq n \leq v_{\text{e}}, \\
n - \left(v_{\text{e}} - v_{\text{s}} + 1 - \left\lfloor \frac{v_{\text{e}} - v_{\text{s}}}{m} \right\rfloor\right), & \text{if } n > v_{\text{e}}.
\end{cases}
\end{align}

\noindent where \( v_{\text{s}} \) and \( v_{\text{e}} \) are the starting and ending position ids of the visual tokens within the global RoPE position id \( n \). \( m \) is the number of visual tokens per frame, and \( \left\lfloor . \right\rfloor \) denotes the floor function, which rounds down to the nearest integer. By scaling the position ids, temporal information is introduced through the adjusted position \(\hat{n}\), defined as:

\begin{align}\label{eq:6}
\hat{n} = n + \gamma \cdot \mathbf{I}_t(n),
\end{align}

\noindent where \(\gamma\) is a scaling factor of constant magnitude. This adjustment ensures that temporal information is effectively incorporated into the original position embedding. For both text and visual tokens, the query and key vectors are updated using the adjusted positions \(\hat{n}\) and \(\hat{m}\):

\begin{align}\label{eq:7}
\mathbf{q}'_{n} &= \mathbf{q}_n e^{i \hat{n} \theta} = \mathbf{q}_n e^{i (n + \gamma \cdot \mathbf{I}_t(n)) \theta} \nonumber , \\
\mathbf{k}'_{m} &= \mathbf{k}_m e^{i \hat{m} \theta} = \mathbf{k}_m e^{i (m + \gamma \cdot \mathbf{I}_t(m)) \theta} ,
\end{align}

Finally, the attention matrix is calculated as follows:

\begin{align}\label{eq:8}
A_{(\hat{n}, \hat{m})} &= \text{Re}[\mathbf{q}'_{n} \mathbf{k}'_{m}{}^*] \nonumber \\
&= \text{Re}[\mathbf{q}_n e^{i (n + \gamma \cdot \mathbf{I}_t(n)) \theta} \mathbf{k}_m^* e^{i (m + \gamma \cdot \mathbf{I}_t(m)) \theta}] \nonumber \\
&= \text{Re}[\mathbf{q}_n \mathbf{k}_m^* e^{i [(n-m) + \gamma (\mathbf{I}_t(n) - \mathbf{I}_t(m))] \theta}]
\end{align}

This formula combines the updated query and key vectors to compute the attention map, incorporating both global positional and temporal information from video frames. By leveraging these aspects, we enhances the MLLM's ability to process and understand the input video comprehensively, resulting in more accurate and contextually appropriate responses.

\begin{figure}[ht]
\centering
   \includegraphics[width=230pt,height=220pt,trim=0 0 0 0,clip]{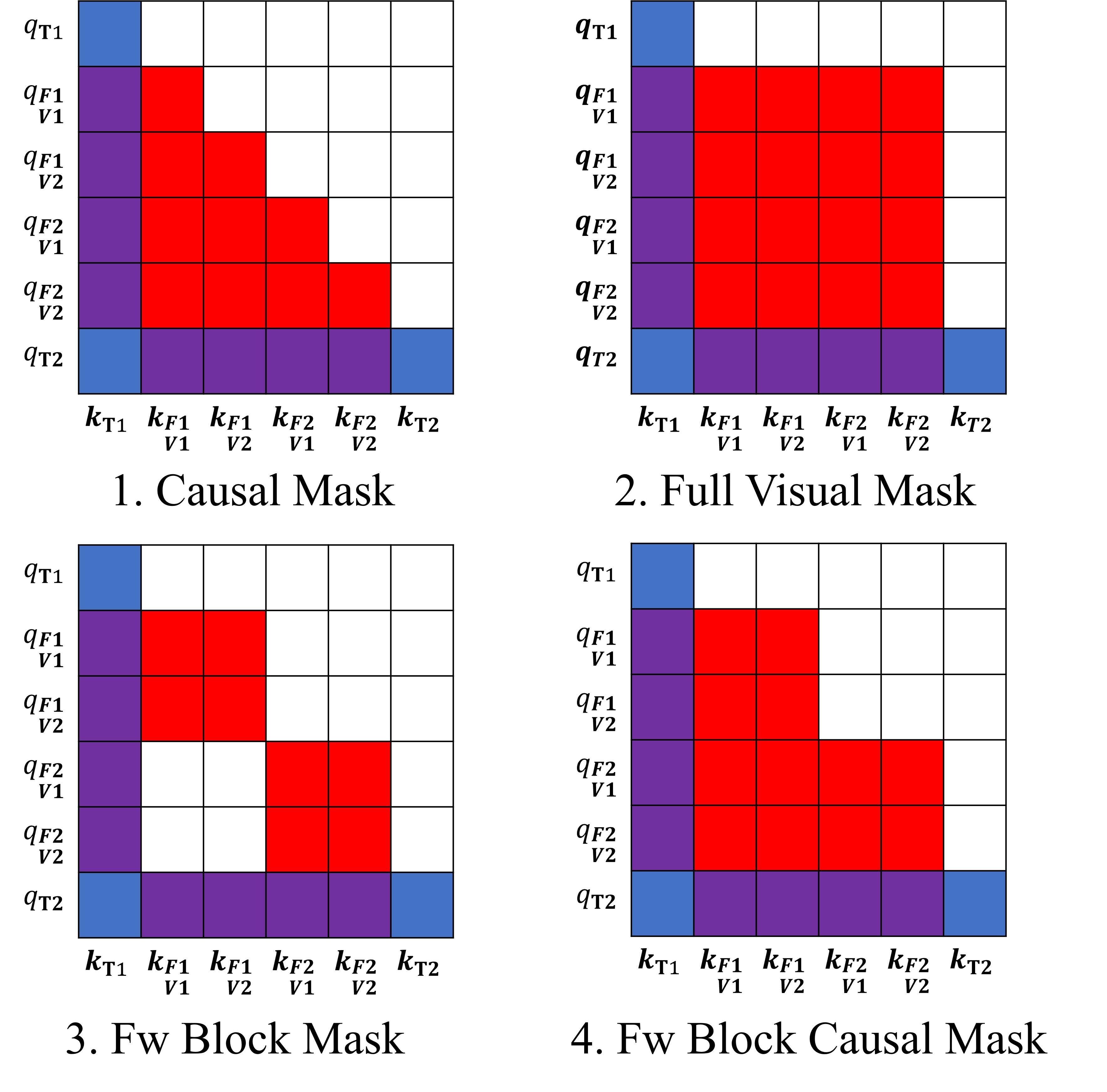}
    \caption{\textbf{Variations of Attention Masks}. To explore attention mechanisms for better interactions, we compare \textbf{Causal Mask}~(1.) with three variants: \textbf{Full Visual Mask}~(2.), \textbf{Frame-wise (Fw) Block Mask}~(3.), and \textbf{Frame-wise (Fw) Block Causal Mask}~(4.). Red indicates pure visual token interactions, blue represents pure text token interactions, and purple denotes interactions between visual and text tokens.}

\label{Framework 3}
\end{figure}

\begin{table*}[ht]
\centering
\scalebox{0.8}{
\begin{tblr}{
  cells = {c},
  cell{1}{1} = {r=2}{},
  cell{1}{2} = {r=2}{},
  cell{1}{3} = {r=2}{},
  cell{1}{4} = {c=2}{},
  cell{1}{6} = {c=2}{},
  cell{1}{8} = {c=2}{},
  cell{1}{10} = {c=2}{},
  cell{1}{12} = {c=6}{},
  hline{1,3, 17, 18} = {-}{},
  hline{2} = {4-17}{},
}
\textbf{Method } & {\textbf{Vision}\\\textbf{Encoder }} & {\textbf{LLM }\\\textbf{Size }} & \textbf{MSVD-QA } &               & \textbf{MSRVTT-QA } &               & \textbf{ActivityNet-QA } &               & \textbf{TGIF-QA } &               & \textbf{Video-ChatGPT } &             &             &             &             &               \\
                 &                                      &                                 & \textbf{Acc.}     & \textbf{Sco.} & \textbf{Acc.}       & \textbf{Sco.} & \textbf{Acc.}            & \textbf{Sco.} & \textbf{Acc.~}    & \textbf{Sco.} & \textbf{CI}             & \textbf{DO} & \textbf{CU} & \textbf{TU} & \textbf{CO} & \textbf{Avg.} \\
Video-LLaMA & CLIP-G & 7B & 51.6 & 2.5 & 29.6 & 1.8 & 12.4 & 1.1 & - & - & 1.96 & 2.18 & 2.16 & 1.82 & 1.79 & 1.98 \\
Video-LLaMA2 & CLIP-L & 7B & 70.9 & 3.8 & - & - & 49.9 & 3.3 & - & - & 3.14 & \textbf{3.08} & 3.69 & 2.56  & \textbf{3.16} & 3.13\\
LLaMA-Adapter & ViT-B & 7B & 54.9 & 3.1 & 43.8 & 2.7 & 34.2 & 2.7 & - & - & 2.03 & 2.32 & 2.30 & 1.98 & 2.15 & 2.16 \\
Video-ChatGPT & ViT-L & 7B & 64.9 & 3.3 & 49.3 & 2.8 & 35.2 & 2.7 & 51.4 & 3.0 & 2.50 & 2.57 & 2.69 & 2.16 & 2.20 & 2.42 \\
Chat-UniVi & ViT-L & 7B & 65.0 & 3.6 & 54.6 & 3.1 & 45.8 & 3.2 & 60.3 & 3.4 & 2.89 & 2.91 & 3.46 & 2.89 & 2.81 & 2.99 \\
MovieChat & CLIP-G & 7B & 75.2 & 3.8 & 52.7 & 2.6 & 45.7 & 3.4 & - & - & 2.76 & 2.93 & 3.01 & 2.24 & 2.42 & 2.67 \\
VideoChat & CLIP-G & 7B & 56.3 & 2.8 & 45.0 & 2.5 & 26.5 & 2.2 & 34.4 & 2.3 & 2.23 & 2.50 & 2.53 & 1.94 & 2.24 & 2.29 \\
VideoChat2 & UMT-L & 7B & 70.0 & 3.9 & 54.1 & 3.3 & 49.1 & 3.3 & - & - & 3.02 & 2.88 & 3.51 & 2.66 & 2.81 & 2.98 \\
Vista-LLaMA & CLIP-G & 7B & 65.3 & 3.6 & 60.5 & 3.3 & 48.3 & 3.3 & - & - & 2.44 & 2.64 & 3.18 & 2.26 & 2.31 & 2.57 \\
LLaMA-VID & CLIP-G & 13B & 70.0 & 3.7 & 58.9 & 3.3 & 47.5 & 3.3 & - & - & 2.96 & 3.00 & 3.53 & 2.46 & 2.51 & 2.89 \\
IG-VLM LLaVA & ViT-L & 7B & \textbf{78.8} & \textbf{4.1} & \underline{63.7} & 3.5 & 54.3 & 3.4 & 73.0 & 4.0 & 3.11 & 2.78 & 3.51 & 2.44 & 3.29 & 3.03 \\
ST-LLM  & BLIP2 & 7B & 74.6 & 3.9 & 63.2 & 3.4 & 50.9 & 3.3 & - & - & 3.23 & \underline{3.05} & \underline{3.74} & \textbf{2.93} & 2.81 & 3.15 \\
PLLaVA & ViT-L & 7B & 76.6 & \textbf{4.1} & 62.0 & \underline{3.5} & 56.3 & \textbf{3.5} & \underline{77.5} & \underline{4.1} & 3.21 & 2.86 & 3.62 & 2.33 & 2.93 & 3.12 \\
GPT-4V  & Unk & Unk & 76.3 & 4.0 & \textbf{63.8} & \underline{3.5} & \textbf{57.0} &\textbf{3.5} & 65.3 & 3.7 & \textbf{3.40} & 2.80 & 3.61 & 2.89 & \underline{3.13} & \underline{3.17} \\
TC-LLaVA & ViT-L & 7B & \textbf{78.8} & \textbf{4.1} & 63.2  & \textbf{3.6} & \underline{56.8}  & \textbf{3.5} & \textbf{78.2} & \textbf{4.2} & \underline{3.25}  & 2.96 &\textbf{3.75} & \underline{2.91} &  3.09 & \textbf{3.19} \\
\end{tblr}}
\caption{Results of video question answering on MSVD-QA, MSRVTT-QA, ActivityNet-QA, TGIF-QA, Video-ChatGPT.}
\label{vqa}
\end{table*}

\subsection{Frame-wise Block Causal Attention Mask}
\label{FWBC}
Another often overlooked key point is the design of attention masks within the transformer layers in large language models. In causal language models like the GPT~\cite{achiam2023gpt} and LLama~\cite{touvron2023llama} series, causal attention masks are employed to ensure that during text aggressive generation, historical token information is not leaked; that is, subsequent tokens can "see" preceding tokens, but preceding tokens cannot "see" subsequent tokens. This design is uniformly applied in such generative models to maintain the unidirectional flow of information, which is crucial for generating coherent and contextually appropriate text. 

Mathematically, the causal attention mask $M \in \mathbb{R}^{T \times T}$ for a sequence of length $T$ is defined as:

\begin{align}\label{eq:9}
M_{ij} =
\begin{cases}
0 & \text{if } i \geq j, \\
-\infty & \text{if } i < j.
\end{cases}
\end{align}

This ensures that each position $i$ only attends to previous positions (including itself), thus implementing the causal attention mechanism. The final attention weights are computed as:

\begin{align}\label{eq:11}
\text{Attention}(Q, K, V) = \text{softmax}\left(\frac{QK^T}{\sqrt{d_k}} + M\right)V,
\end{align}

\noindent where $Q$ is the query vectors, $K$ is the key vectors, $V$ is the value vectors, $d_k$ is the dimension of the key vectors, and $M$ is the causal attention mask.

However, for multimodal information involving both visual and textual inputs, the visual modality is only used as a conditional input to the language model. During the unidirectional decoding process of the language model, this design weakens the bidirectional attention interactions obtained from the visual encoder, reducing them to unidirectional attention interactions. To explore the impact of different attention masks, we design three distinct attention masks to enhance and investigate better interactions within visual tokens and between visual and text tokens, as illustrated in Figure~\ref{Framework 3}.

Firstly, the \textbf{Full Visual Mask} modifies the causal attention mask to enable more extensive interactions among visual tokens across different frames. This mask can be represented as follows:
\[
M_{ij}^{\text{Full Visual}} =
\begin{cases}
0 & \text{if } i \geq j \text{ or } i,j \text{ are visual tokens}, \\
-\infty & \text{otherwise}.
\end{cases}
\]

Secondly is \textbf{Frame-wise Block Mask}, which limits the attention to adjacent visual tokens within the same frame. This is defined as follows:

\[
M_{ij}^{\text{Fw Block}} =
\begin{cases}
0 & \text{if } i \geq j \text{ and } i, j \text{ within the same frame}, \\
-\infty & \text{otherwise}.
\end{cases}
\]

Finally, we proposed \textbf{Frame-wise Block Causal Attention Mask} (FwBC), which combines the characteristics of the previous causal and block visual attention masks by incorporating broader visual token interactions within the frame while maintaining causal inference mode across video frames. This can be presented as:

\[
M_{ij}^{\text{FwBC}} =
\begin{cases}
0 & \text{if } i \geq j \text{ or } i, j \text{ within the same frame}, \\
-\infty & \text{otherwise}.
\end{cases}
\]

By adjusting these masks, we aim to achieve a better balance between visual and textual information integration, enabling MLLMs to distinguish and process both video and text modalities more effectively while enhancing the spatiotemporal global attention to the most critical visual modality information for video understanding tasks. Finally, we utilized ablation experiments to select the Frame-wise Block causal Attention Mask for constructing TC-LLaVA.

\begin{table*}[ht]
\centering
\scalebox{0.64}{
\begin{tblr}{
  cells = {c},
  hline{1,12} = {-}{0.08em},
  hline{2,10-11} = {-}{0.05em},
}
\textbf{Method}     & {\textbf{Vision }\\\textbf{Encoder}} & {\textbf{LLM }\\\textbf{Size}} & \textbf{AS}   & \textbf{AP}   & \textbf{AA}   & \textbf{FA}   & \textbf{UA}   & \textbf{OE}   & \textbf{OI}   & \textbf{OS}   & \textbf{MD}   & \textbf{AL}   & \textbf{ST}   & \textbf{AC}   & \textbf{MC}   & \textbf{MA}   & \textbf{SC}   & \textbf{FP} & \textbf{CO}   & \textbf{EN}   & \textbf{ER}   & \textbf{CI}   & \textbf{Avg}. \\
Video-LLLaMA        & CLIP-G                               & 7B                             & 27.5          & 25.5          & 51.0          & 29.0          & 39.0          & 48.0          & 40.5          & 38.0          & 22.5          & 22.5          & 43.0          & 34.0          & 22.5          & 32.5          & 45.5          & 32.5        & 40.0          & 30.0          & 21.0          & 37.0          & 34.1          \\
LLLaMA-Adapter      & ViT-B                                & 7B                             & 23.0          & 28.0          & 51.0          & 30.0          & 35.0          & 35.0          & 33.5          & 33.5          & 21.5          & 21.5          & 36.0          & 29.0          & 31.5          & 32.5          & 44.5          & 31.5        & 31.5          & 22.5          & 28.0          & 32.0          & 31.7          \\
Video-ChatGPT       & ViT-L                                & 7B                             & 23.5          & 26.0          & 62.0          & 22.5          & 26.5          & 54.0          & 28.0          & 30.0          & 23.0          & 20.0          & 31.0          & 30.0          & 25.5          & 39.5          & \underline{48.5}          & 29.0        & 40.0          & 25.0          & 26.0          & 35.0          & 32.7          \\
VideoChat           & CLIP-G                               & 7B                             & 33.5          & 26.5          & 56.0          & 33.5          & 40.5          & 53.0          & 40.5          & 30.0          & 25.5          & 27.0          & 48.5          & 35.0          & 20.5          & 42.5          & 46.0          & 26.5        & 41.0          & 23.5          & 23.5          & 36.0          & 35.5          \\
VideoChat2          & UMT-L                                & 7B                             & \underline{66.0}          & 47.5          & \underline{83.5}          & \textbf{49.5} & 60.0          & 58.0          & \underline{71.5}          & \textbf{42.5} & 23.0          & 23.0          & \textbf{88.5} & \underline{39.0}          & 42.0          & 58.5          & 44.0          & \textbf{49.0}        & 36.5          & \textbf{35.0} & 40.5          & \underline{65.5}          & 51.1          \\
ST-LLM              & BLIP2                                & 7B                             & \underline{66.0}          & 53.5          & \textbf{84.0} & 44.0          & 58.5          & \underline{80.5}          & \textbf{73.5} & \underline{38.5}          & \textbf{42.5} & 31.0          & \underline{86.5}          & 36.5          & \underline{56.5}          & \underline{78.5}          & 43.0          & 44.5        & 46.5          & \underline{34.5}          & 41.5          & 58.5          & \underline{54.9}          \\
PLLaVA              & ViT-L                                & 7B                             & 58.0          & 49.0          & 55.5          & 41.0          & \underline{61.0} & 56.0          & 61.0          & 36.0          & 23.5          & 26.0          & 82.0          & \textbf{39.5} & 42.0          & 52.0          & 45.0          & 42.0        & \underline{53.5}          & 30.5          & \underline{48.0} & 31.0          & 46.6          \\
GPT-4V              & Unk                               & Unk                            & 55.5          & \textbf{63.5}          & 72.0          & \underline{46.5}          & \textbf{73.5}          & 18.5          & 59.0          & 29.5          & 12.0          & \textbf{40.5}          & 83.5          & \underline{39.0}          & 12.0          & 22.5          & 45.0          & \underline{47.5}        & 52.0          & 31.0          & \textbf{59.0}          & 11.0          & 43.5          \\
TC-LLaVA            & ViT-L                                & 7B                             & \textbf{71.5} & \underline{56.5} & 67.5          & 44.5          & 59.5          & \textbf{84.0} & 70.0          & 37.0          & \underline{39.5}          & \underline{39.5} & 85.5          & 35.5          & \textbf{59.5} & \textbf{83.5} & \textbf{53.5} & 42.0        & \textbf{54.0} & 32.0          & 47.0          & \textbf{70.0} & \textbf{56.6} \\
\end{tblr}}
\caption{Results on MVBench multi-choice question answering.}
\label{tab:mvbench_results}
\end{table*}

\section{Experiments}
\subsection{Experimental Setup}
\textbf{Instruction Tuning Datasets.} In alignment with the instruction tuning setting outlined in VideoChat2~\cite{2023videochat}, which integrates data for a variety of video understanding tasks, we utilized an extensive and diverse collection of datasets. Specifically,  these include 27k conversation videos from VideoChat~\cite{li2023videochat} and Video-ChatGPT~\cite{maaz2023video}, 80k classification task samples from Kinetics~\cite{kay2017kinetics} and SthSthV2~\cite{goyal2017something}, 450k captioned data from Webvid~\cite{bain2021frozen}, YouCook2~\cite{zhou2018towards}, TextVR~\cite{wu2023large}, and VideoChat, 117 reasoning data samples from NextQA~\cite{xiao2021next}, CLEVRER~\cite{yi2019clevrer}, and 109,000 annotated question answering samples from Webvid, TGIF~\cite{li2016tgif}, and Ego4D~\cite{grauman2022ego4d}. In total, we employed 783k video instruction data samples for conducting supervised finetuning (SFT) our TC-LLaVA.\\

\noindent\textbf{Evaluation Benchmarks.} The performance of our trained TC-LLaVA model is assessed using a series of video understanding benchmarks, specifically targeting open-ended Video Question Answer (VideoQA) tasks. These benchmarks include MSVD-QA~\cite{xu2017video}, MSRVTT-QA~\cite{xu2016msr}, Activity-QA~\cite{caba2015activitynet}, and TGIF-QA~\cite{li2016tgif}, where responses generally consist of single-word answers. The accuracy (with true/false answers) and quality (scored from 0 to 5) of the models' responses are evaluated using GPT-3.5~\cite{chatgpt}. Moreover, we employ the Video-based Generative Performance benchmark (VCG Score), as introduced by VideoChatGPT~\cite{maaz2023video}. This benchmark requires longer answers and evaluates five key aspects of video understanding: Correctness of Information (CI), Detail Orientation (DO), Context Understanding (CU), Temporal Understanding (TU), and Consistency (CO). The generative performance is also assessed using the GPT-3.5 model. In addition, we evaluate TC-LLaVA on the multi-choice Question Answering benchmark, MVBench~\cite{li2023mvbench}, which consists of 20 tasks that demand nuanced temporal comprehension of videos.\\

\noindent\textbf{Implementation Details} Initialized from the image-pretrained MLLM LLaVA-Next~\cite{zhang2024llavanextvideo}, which is based on the Vicuna-7B-v1.5~\cite{zheng2024judging}, our TC-LLaVA 7B conduct further video instruction supervised finetuning (SFT) and evaluation on the datasets mentioned above. Following the experimental settings in ~\cite{xu2024pllava}, we uniformly sample 16 frames from the raw video as input and use global average pooling to downsample the visual features from a shape of \(\ 24*24*d \)\ to \(\ 12*12*d\)\, where d represents the input feature dimension of the LLM part. During the SFT stage, we employ a batch size of 128 and a learning rate of 2e-5, utilizing a cosine scheduler and a warmup ratio of 0.03. All reported results are evaluated on models trained for 7k steps on 8 NVIDIA A100 GPU. For evaluation, we use the GPT-3.5-turbo-0125 model across benchmarks that require additional scoring or assessment.

\subsection{Comparison with SOTA}
In this section, we compare our TC-LLaVA with recent advanced works, including Video-LLaMA~\cite{zhang2023video}, LLaMA-Adapter~\cite{zhang2023llama}, Video-ChatGPT~\cite{maaz2023video}, Chat-UniVi~\cite{jin2024chat}, MovieChat~\cite{su2020moviechats}, VideoChat~\cite{2023videochat}, VideoChat2~\cite{li2023mvbench}, Vista-LLaMA~\cite{ma2023vista}, LLaMA-VID~\cite{li2023llama}, IG-VLM LLaVA~\cite{kim2024image}, ST-LLM~\cite{liu2024st}, PLLaVA~\cite{xu2024pllava}, and GPT-4V~\cite{achiam2023gpt}, across various video understanding benchmarks. The best performance is indicated in \textbf{bold}, and the second-best results are indicated with \underline{underlining}. As shown in Table~\ref{vqa}, our TC-LLaVA achieves a new state-of-the-art performance across MSVD-QA, TGIF-QA, and Video-ChatGPT, surpassing GPT-4V by 2.5\%, 7.9\%, and 0.02\%, respectively. Additionally, our TC-LLaVA achieves the best performance across video question-answering benchmarks on the Score metric.  Compared to the latest work PLLaVA, which is also initialized from LLaVA-Next and continues using original causal attention mask and RoPE, our TC-LLaVA outperforms it across all five evaluation benchmarks, demonstrating the effectiveness of our proposed methods. 

Furthermore, we evaluate TC-LLaVA on MVbench, a multiple-choice video question answering benchmark, focusing on questions that require comprehensive understanding of the entire video. As shown in Table~\ref{tab:mvbench_results},  TC-LLaVA achieves state-of-the-art performance in the average MVbench score. Specifically, for time-related tasks such as Action Sequence (AS), Object Existence (OE), Moving Count (MC), Moving Attribute (MA), State Change (SC), Character Order (CO), and Counterfactual Inference (CI), TC-LLaVA demonstrates a significant performance margin of at least 0.5\% over other open-source models. Even when compared to GPT-4V, we maintain an edge in average performance across all 20 tasks by 13.1\%.

\begin{figure}[ht]
\centering
   \includegraphics[width=1.0\linewidth]{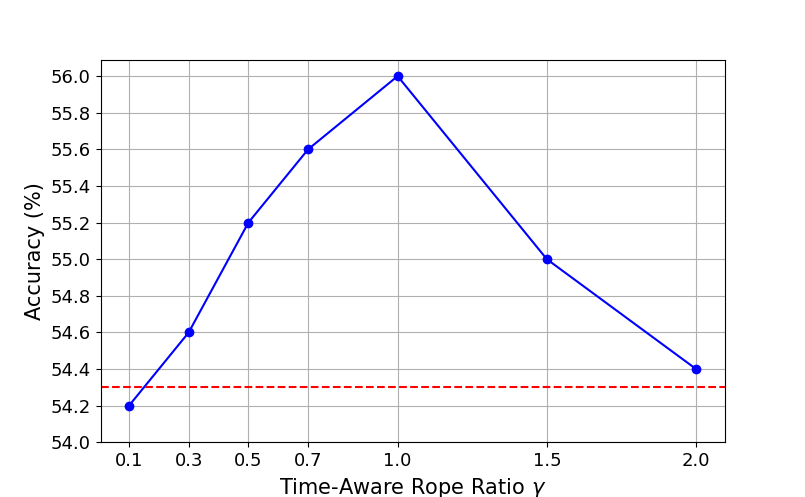}
   \caption{Different ratio $\gamma$ settings of Time-Aware RoPE on MVbench. The red dashed line in the figure represents the baseline performance, which is the performance without adding the time-aware rope. The blue line shows the performance variations of the model under different ratio settings.}
\label{Ablation1}
\end{figure}

\subsection{Ablation Studies}
In this subsection, we conduct ablation studies to assess the impact of key components. Specifically, we examine the manual ratio settings $\gamma$ of Time-Aware Dual RoPE and other designs of the attention mask, beyond the proposed Frame-wise Block Causal Mask as shown in Figure~\ref{Framework 3}. For these studies, we use the basic settings as a combination of both the original RoPE and Causal Attention Mask, while keeping the previously mentioned training settings. The evaluation is performed on MVbench. Finally, we present a visualized heatmap comparing the attention weights of our TC-Attention mechanism to the vanilla attention.

\subsubsection{Time-Aware RoPE Ablation} 
Firstly, Maintaining global Rotary Position Embedding (RoPE) is crucial for preserving the global positional relationships between tokens. LLaVA treated each token in an image as having an independent position. When transitioning from image to video understanding tasks, we aim to retain the characteristics of these pre-trained weights while introducing time-aware RoPE to incorporate temporal information. If we entirely abandon the use of RoPE, it could result in a partial loss of the capabilities encoded in the pre-trained LLMs, ultimately affecting the final performance. 

Secondly, RoPE employs a rotational invariant mechanism, which contrasts with the linear and fixed positional embedding schemes of absolute and learnable embeddings. These inherent differences can hinder RoPE's effective scalability when integrating it with other positional encoding techniques, potentially resulting in suboptimal performance or conflicting representations.

Finally, we explore the impact of the hyperparameter $\gamma$ in Time-Aware Dual RoPE. As shown in Figure~\ref{Ablation1}, we evaluate TC-LLaVA on MVbench by setting the manual ratio $\gamma$ across [0.1, 0.3, 0.5, 0.7, 1.0, 1.5, 2.0]. Compared to the baseline setting, which uses a single global RoPE (indicated by the red dashed line), introducing our Time-Aware RoPE increases performance, particularly when $\gamma$ is close to 1.0, achieving the best performance at 56.0\%. However, further increasing $\gamma$ slightly reduces the final performance. We think this occurs because increasing $\gamma$ too much might distort the original global position relationships encoded by the original RoPE, leading to suboptimal integration of spatial and temporal information. In the end, we choose $\gamma$ as 1.0 for TC-LLaVA's experimental setting across the entire paper. 

\begin{table}
\centering
\begin{tabular}{c|c|c} 
\hline
\textbf{Methods} & \textbf{Mvbench} & \textbf{VCGbench}  \\ 
\hline
Baseline            & 54.3                & 3.09                   \\ 
Full Visual Mask  & 53.8                & 3.04                   \\ 
Fw. Block Mask        & 54.0                & 3.08                   \\ 
Fw. Block causal Mask & \textbf{55.9}      & \textbf{3.13}          \\
\hline
Just Time-Aware RoPE     & 54.2                & 3.11                   \\ 
Time-Aware Dual RoPE       & \textbf{56.0}                & \textbf{3.15}   \\ 
\hline
TC-LLaVA & \textbf{56.6}                & \textbf{3.19}                   \\
\hline
\end{tabular}
\caption{Ablation Study of Different Method Settings. The baseline settings use the original RoPE and Causal Mask.}
\label{Ablation2}
\end{table}

\begin{figure}[h]
\centering
   \includegraphics[width=1.0\linewidth,trim=0 18 0 0,clip]{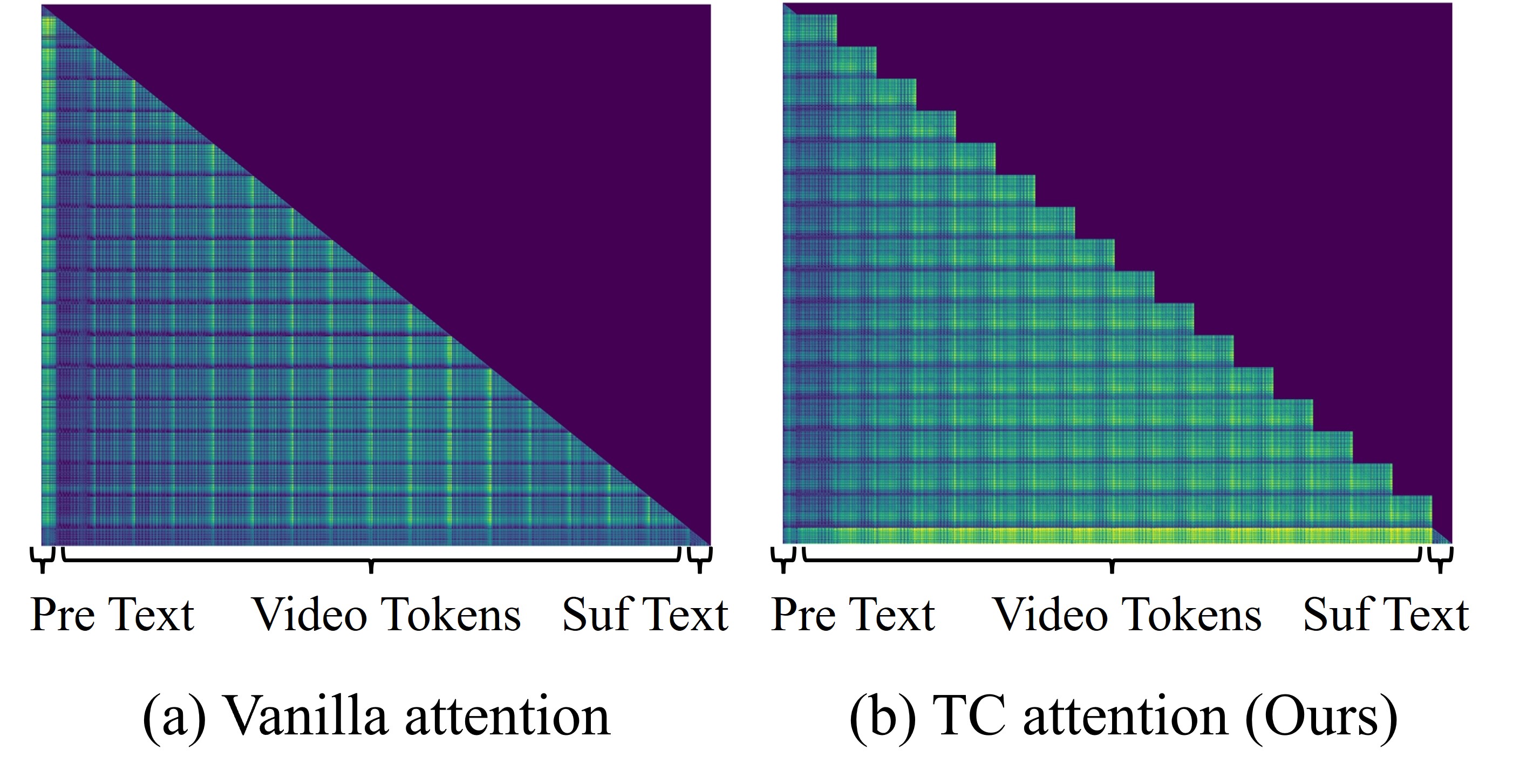}
   \caption{Comparison of attention weights with corresponding attention mask for Vanilla attention (a) and TC attention (b). Lighter colors represent higher weights.}
\label{Attention}
\end{figure}

\subsubsection{Attention Mask and Combination Ablation}
We further explore other attention mask variances mentioned above. As shown in Table~\ref{Ablation2}, using Full Visual and Frame-wise (Fw.) Block Masks enhances visual token interactions within frames but weakens or sacrifices causal relationships. This is crucial for video understanding, as future frames should be able to reference previous frames, but previous frames should avoid seeing future frames, similar to the way text sequences are handled in autoregressive generation. Our Fw. Block Causal Mask achieves better performance by considering both enhancing visual interactions and preserving the causal relationships between tokens. When combined with Time-Aware Dual RoPE, our TC-LLaVA demonstrates superior performance, scoring 56.6\% on MVbench and 3.19 on VCGbench. This combination improves the Attention Module, the core component of the LLM, resulting in a more comprehensive and effective video understanding model.

\begin{table}
\centering
\label{inat}
\begin{tblr}{
  column{2} = {c},
  column{3} = {c},
  cell{3}{2} = {c=2}{},
  cell{4}{2} = {c=2}{},
  hline{1-2,6,9} = {-}{},
}
Method                  & MVbench      & VCGbench  \\
RoPE~(baseline)         & 54.3         & 3.09      \\
Time APE                     & Not Converge &           \\
Time RPE                     & Not Converge &           \\
Time RoPE               & 54.2         & 3.11      \\
RoPE + Time APE              & 50.2      & 3.02   \\
RoPE + Time RPE              & 52.7      &  3.08  \\
RoPE + Time RoPE (Ours) & \textbf{56.0}    & \textbf{3.15} \\
\end{tblr}
\caption{The Ablation Studies on Different PE Settings.}
\label{RoPE AB}
\end{table}

\section{Other Time-Aware Position Embedding}
To further verify the effectiveness of our Time-Aware Dual RoPE, we conducted the ablation study presented in Table~\ref{RoPE AB}, which demonstrates the comparative performance of different positional encoding (PE) settings on two benchmarks, MVbench~\cite{li2023mvbench} and VCGbench~\cite{maaz2023video}. The experimental settings used are consistent with those outlined in the paper. The baseline setting RoPE achieves a score of 54.3 on MVbench and 3.09 on VCGbench. However, when introducing Time Absolute Position Encoding (APE) and Time Relative Position Encoding (RPE) individually, the models fail to converge. This issue arises because the pre-trained language model (LLM) is based on RoPE, and the inherent differences between APE, RPE, and RoPE result in significant alterations to the inter-layer features learned during pre-training. Consequently, these discrepancies cause instability in the training loss, leading to fluctuations that hinder the model's ability to converge effectively. Therefore, the effective approach to incorporating temporal information into a pre-trained LLM is through Time RoPE, as it minimizes conflicts with the pre-existing model configurations. By aligning more closely with the RoPE framework used during pre-training, Time RoPE ensures a smoother integration of temporal features, thereby reducing instability during supervised finetuning (SFT) stage. By combining with the original RoPE, our proposed RoPE + Time RoPE (Time-Aware Dual RoPE) achieves further improvement and outperforms all other configurations, with enhanced scores of 56.0 on MVbench and 3.15 on VCGbench, demonstrating the effectiveness of our approach in PE setting methods and leveraging the spatial-temporal positional information.

\begin{table}
\centering
\label{inat}
\begin{tblr}{
  column{2} = {c},
  column{3} = {c},
  hline{1-2,4, 6,8} = {-}{},
}  TC-LLaVA                & MVbench   & VCGbench  \\
Vicuna-7B-v1.5             & 54.3      & 3.09   \\
\textbf{+ TC-Attention}    & 56.6 (+2.3)     & 3.19 (+0.1)  \\
Llama3-8B-Instruct         & 56.2      & 3.12   \\
\textbf{+ TC-Attention}    & 57.5 (+1.0)    & 3.18  (+0.06) \\
Mistral-7B-Instruct-v0.2   & 53.7      & 3.06 \\
\textbf{+ TC-Attention}    & 55.8 (+2.1)     & 3.15 (+0.09)\\
\end{tblr}
\caption{More Studies on Different Base Model.}
\label{base models}
\end{table}

\section{TC-Attention on Different Base Model} 
To further assess the generalizability and robustness of our TC-Attention mechanism, we extended its application beyond Vicuna-7B-v1.5 to include other pre-trained LLM base models, specifically Llama3-8B-Instruct~\cite{llama3modelcard} and Mistral-7B-Instruct-v0.2~\cite{jiang2023mistral}.  Each model was initialized from the pre-trained LLaVa-Next models~\cite{li2024llavanext-strong}\footnote{https://huggingface.co/collections/llava-hf/llava-next-65f75c4afac77fd37dbbe6cf}, with all subsequent fine-tuning experiments conducted under consistent SFT settings. The performance of the fine-tuned models was evaluated on two benchmarks, MVbench~\cite{li2023mvbench} and VCGbench~\cite{maaz2023video}, with the results summarized in Table \ref{base models}. Across different base models, the introduction of TC-Attention led to measurable improvements in performance on both benchmarks. These results underscore the efficacy of TC-Attention, demonstrating its ability to enhance the performance of diverse base models. The consistent gains observed across different architectures not only validate the adaptability of TC-Attention but also highlight its potential as a valuable component in optimizing the performance of large language models on complex tasks.

\subsubsection{Attention Visualization}
Finally, we illustrate the attention weights of both our TC-Attention and Vanilla Attention. For this experiment, we compare the video-finetuned LLaVA and TC-LLaVA by inputting the same video test samples and visualizing the average attention weights of different heads in the final decoding layer of the LLM. In the visualization of Figure~\ref{Attention}, brighter colors represent higher weights while the darker color represent lower weights. The attention weights assigned to visual tokens are markedly more comprehensive and greater in our TC-Attention. This indicates that, unlike Vanilla Attention, which only focuses on the last few visual tokens of each frame, our TC-Attention attends to every visual token within and across frames. Additionally, the proposed TC-Attention assigns greater attention weight to subsequent text (user input), resulting in a considerably more substantial impact of visual tokens on language tokens. This demonstrates the effectiveness of TC-Attention in integrating visual and textual information, enhancing the model's overall understanding and performance.

\section{Conclusion}
In this work, we present TC-LLaVA, rethinking the attention design in large language models (LLM) for video tasks. We introduce two core components to achieve this: Temporal-Aware Dual RoPE, incorporating temporal information into the attention module while maintaining the global position information between visual and text tokens, and Frame-wise Block Causal Attention Mask, enhancing the interaction of visual tokens within frames while preserving causal relationships across video frames. By conducting simple supervised finetuning (SFT) on video-related instruction datasets, our TC-LLaVA achieves new state-of-the-art performance across various video understanding benchmarks, showcasing the effectiveness of these methods. As LLMs continue to scale up, their powerful performance has led to the negligence of some design details. We hope our work encourages researchers to rethink these design aspects.


\newpage
{\small
\bibliographystyle{ieee_fullname}
\bibliography{egbib}

\begin{thebibliography}{10}\itemsep=-1pt

\bibitem{achiam2023gpt}
Josh Achiam, Steven Adler, Sandhini Agarwal, Lama Ahmad, Ilge Akkaya, Florencia~Leoni Aleman, Diogo Almeida, Janko Altenschmidt, Sam Altman, Shyamal Anadkat, et~al.
\newblock Gpt-4 technical report.
\newblock {\em arXiv preprint arXiv:2303.08774}, 2023.

\bibitem{llama3modelcard}
AI@Meta.
\newblock Llama 3 model card.
\newblock 2024.

\bibitem{bahdanau2014neural}
Dzmitry Bahdanau, Kyunghyun Cho, and Yoshua Bengio.
\newblock Neural machine translation by jointly learning to align and translate.
\newblock {\em arXiv preprint arXiv:1409.0473}, 2014.

\bibitem{bain2021frozen}
Max Bain, Arsha Nagrani, G{\"u}l Varol, and Andrew Zisserman.
\newblock Frozen in time: A joint video and image encoder for end-to-end retrieval.
\newblock In {\em Proceedings of the IEEE/CVF international conference on computer vision}, pages 1728--1738, 2021.

\bibitem{caba2015activitynet}
Fabian Caba~Heilbron, Victor Escorcia, Bernard Ghanem, and Juan Carlos~Niebles.
\newblock Activitynet: A large-scale video benchmark for human activity understanding.
\newblock In {\em Proceedings of the ieee conference on computer vision and pattern recognition}, pages 961--970, 2015.

\bibitem{chen2023sharegpt4v}
Lin Chen, Jisong Li, Xiaoyi Dong, Pan Zhang, Conghui He, Jiaqi Wang, Feng Zhao, and Dahua Lin.
\newblock Sharegpt4v: Improving large multi-modal models with better captions.
\newblock {\em arXiv preprint arXiv:2311.12793}, 2023.

\bibitem{cheng2024videollama}
Zesen Cheng, Sicong Leng, Hang Zhang, Yifei Xin, Xin Li, Guanzheng Chen, Yongxin Zhu, Wenqi Zhang, Ziyang Luo, Deli Zhao, et~al.
\newblock Videollama 2: Advancing spatial-temporal modeling and audio understanding in video-llms.
\newblock {\em arXiv preprint arXiv:2406.07476}, 2024.

\bibitem{dosovitskiy2020image}
Alexey Dosovitskiy, Lucas Beyer, Alexander Kolesnikov, Dirk Weissenborn, Xiaohua Zhai, Thomas Unterthiner, Mostafa Dehghani, Matthias Minderer, Georg Heigold, Sylvain Gelly, et~al.
\newblock An image is worth 16x16 words: Transformers for image recognition at scale.
\newblock {\em arXiv preprint arXiv:2010.11929}, 2020.

\bibitem{du2021glm}
Zhengxiao Du, Yujie Qian, Xiao Liu, Ming Ding, Jiezhong Qiu, Zhilin Yang, and Jie Tang.
\newblock Glm: General language model pretraining with autoregressive blank infilling.
\newblock {\em arXiv preprint arXiv:2103.10360}, 2021.

\bibitem{goyal2017something}
Raghav Goyal, Samira Ebrahimi~Kahou, Vincent Michalski, Joanna Materzynska, Susanne Westphal, Heuna Kim, Valentin Haenel, Ingo Fruend, Peter Yianilos, Moritz Mueller-Freitag, et~al.
\newblock The" something something" video database for learning and evaluating visual common sense.
\newblock In {\em Proceedings of the IEEE international conference on computer vision}, pages 5842--5850, 2017.

\bibitem{grauman2022ego4d}
Kristen Grauman, Andrew Westbury, Eugene Byrne, Zachary Chavis, Antonino Furnari, Rohit Girdhar, Jackson Hamburger, Hao Jiang, Miao Liu, Xingyu Liu, et~al.
\newblock Ego4d: Around the world in 3,000 hours of egocentric video.
\newblock In {\em Proceedings of the IEEE/CVF Conference on Computer Vision and Pattern Recognition}, pages 18995--19012, 2022.

\bibitem{jiang2023mistral}
Albert~Q Jiang, Alexandre Sablayrolles, Arthur Mensch, Chris Bamford, Devendra~Singh Chaplot, Diego de~las Casas, Florian Bressand, Gianna Lengyel, Guillaume Lample, Lucile Saulnier, et~al.
\newblock Mistral 7b.
\newblock {\em arXiv preprint arXiv:2310.06825}, 2023.

\bibitem{jin2024chat}
Peng Jin, Ryuichi Takanobu, Wancai Zhang, Xiaochun Cao, and Li Yuan.
\newblock Chat-univi: Unified visual representation empowers large language models with image and video understanding.
\newblock In {\em Proceedings of the IEEE/CVF Conference on Computer Vision and Pattern Recognition}, pages 13700--13710, 2024.

\bibitem{kay2017kinetics}
Will Kay, Joao Carreira, Karen Simonyan, Brian Zhang, Chloe Hillier, Sudheendra Vijayanarasimhan, Fabio Viola, Tim Green, Trevor Back, Paul Natsev, et~al.
\newblock The kinetics human action video dataset.
\newblock {\em arXiv preprint arXiv:1705.06950}, 2017.

\bibitem{kim2024image}
Wonkyun Kim, Changin Choi, Wonseok Lee, and Wonjong Rhee.
\newblock An image grid can be worth a video: Zero-shot video question answering using a vlm.
\newblock {\em arXiv preprint arXiv:2403.18406}, 2024.

\bibitem{li2024llavanext-strong}
Bo Li, Kaichen Zhang, Hao Zhang, Dong Guo, Renrui Zhang, Feng Li, Yuanhan Zhang, Ziwei Liu, and Chunyuan Li.
\newblock Llava-next: Stronger llms supercharge multimodal capabilities in the wild, May 2024.

\bibitem{li2023blip}
Junnan Li, Dongxu Li, Silvio Savarese, and Steven Hoi.
\newblock Blip-2: Bootstrapping language-image pre-training with frozen image encoders and large language models.
\newblock In {\em International conference on machine learning}, pages 19730--19742. PMLR, 2023.

\bibitem{2023videochat}
Kunchang Li, Yinan He, Yi Wang, Yizhuo Li, Wenhai Wang, Ping Luo, Yali Wang, Limin Wang, and Yu Qiao.
\newblock Videochat: Chat-centric video understanding.
\newblock {\em arXiv preprint arXiv:2305.06355}, 2023.

\bibitem{li2023videochat}
KunChang Li, Yinan He, Yi Wang, Yizhuo Li, Wenhai Wang, Ping Luo, Yali Wang, Limin Wang, and Yu Qiao.
\newblock Videochat: Chat-centric video understanding.
\newblock {\em arXiv preprint arXiv:2305.06355}, 2023.

\bibitem{li2023mvbench}
Kunchang Li, Yali Wang, Yinan He, Yizhuo Li, Yi Wang, Yi Liu, Zun Wang, Jilan Xu, Guo Chen, Ping Luo, Limin Wang, and Yu Qiao.
\newblock Mvbench: A comprehensive multi-modal video understanding benchmark, 2023.

\bibitem{li2016tgif}
Yuncheng Li, Yale Song, Liangliang Cao, Joel Tetreault, Larry Goldberg, Alejandro Jaimes, and Jiebo Luo.
\newblock Tgif: A new dataset and benchmark on animated gif description.
\newblock In {\em Proceedings of the IEEE Conference on Computer Vision and Pattern Recognition}, pages 4641--4650, 2016.

\bibitem{li2023llama}
Yanwei Li, Chengyao Wang, and Jiaya Jia.
\newblock Llama-vid: An image is worth 2 tokens in large language models.
\newblock {\em arXiv preprint arXiv:2311.17043}, 2023.

\bibitem{lin2014microsoft}
Tsung-Yi Lin, Michael Maire, Serge Belongie, James Hays, Pietro Perona, Deva Ramanan, Piotr Doll{\'a}r, and C~Lawrence Zitnick.
\newblock Microsoft coco: Common objects in context.
\newblock In {\em Computer Vision--ECCV 2014: 13th European Conference, Zurich, Switzerland, September 6-12, 2014, Proceedings, Part V 13}, pages 740--755. Springer, 2014.

\bibitem{liu2024improved}
Haotian Liu, Chunyuan Li, Yuheng Li, and Yong~Jae Lee.
\newblock Improved baselines with visual instruction tuning.
\newblock In {\em Proceedings of the IEEE/CVF Conference on Computer Vision and Pattern Recognition}, pages 26296--26306, 2024.

\bibitem{liu2024visual}
Haotian Liu, Chunyuan Li, Qingyang Wu, and Yong~Jae Lee.
\newblock Visual instruction tuning.
\newblock {\em Advances in neural information processing systems}, 36, 2024.

\bibitem{liu2024st}
Ruyang Liu, Chen Li, Haoran Tang, Yixiao Ge, Ying Shan, and Ge Li.
\newblock St-llm: Large language models are effective temporal learners.
\newblock {\em arXiv preprint arXiv:2404.00308}, 2024.

\bibitem{liu2022swin}
Ze Liu, Han Hu, Yutong Lin, Zhuliang Yao, Zhenda Xie, Yixuan Wei, Jia Ning, Yue Cao, Zheng Zhang, Li Dong, et~al.
\newblock Swin transformer v2: Scaling up capacity and resolution.
\newblock In {\em Proceedings of the IEEE/CVF conference on computer vision and pattern recognition}, pages 12009--12019, 2022.

\bibitem{liu2021swin}
Ze Liu, Yutong Lin, Yue Cao, Han Hu, Yixuan Wei, Zheng Zhang, Stephen Lin, and Baining Guo.
\newblock Swin transformer: Hierarchical vision transformer using shifted windows.
\newblock In {\em Proceedings of the IEEE/CVF international conference on computer vision}, pages 10012--10022, 2021.

\bibitem{ma2023vista}
Fan Ma, Xiaojie Jin, Heng Wang, Yuchen Xian, Jiashi Feng, and Yi Yang.
\newblock Vista-llama: Reliable video narrator via equal distance to visual tokens.
\newblock {\em arXiv preprint arXiv:2312.08870}, 2023.

\bibitem{maaz2023video}
Muhammad Maaz, Hanoona Rasheed, Salman Khan, and Fahad~Shahbaz Khan.
\newblock Video-chatgpt: Towards detailed video understanding via large vision and language models.
\newblock {\em arXiv preprint arXiv:2306.05424}, 2023.

\bibitem{chatgpt}
OpenAI.
\newblock Chatgpt.
\newblock In {\em https://openai.com/blog/chatgpt}, 2023.

\bibitem{schuhmann2022laion}
Christoph Schuhmann, Romain Beaumont, Richard Vencu, Cade Gordon, Ross Wightman, Mehdi Cherti, Theo Coombes, Aarush Katta, Clayton Mullis, Mitchell Wortsman, et~al.
\newblock Laion-5b: An open large-scale dataset for training next generation image-text models.
\newblock {\em Advances in Neural Information Processing Systems}, 35:25278--25294, 2022.

\bibitem{shaw2018self}
Peter Shaw, Jakob Uszkoreit, and Ashish Vaswani.
\newblock Self-attention with relative position representations.
\newblock {\em arXiv preprint arXiv:1803.02155}, 2018.

\bibitem{su2020moviechats}
Hui Su, Xiaoyu Shen, Zhou Xiao, Zheng Zhang, Ernie Chang, Cheng Zhang, Cheng Niu, and Jie Zhou.
\newblock Moviechats: Chat like humans in a closed domain.
\newblock In {\em Proceedings of the 2020 conference on empirical methods in natural language processing (EMNLP)}, pages 6605--6619, 2020.

\bibitem{su2024roformer}
Jianlin Su, Murtadha Ahmed, Yu Lu, Shengfeng Pan, Wen Bo, and Yunfeng Liu.
\newblock Roformer: Enhanced transformer with rotary position embedding.
\newblock {\em Neurocomputing}, 568:127063, 2024.

\bibitem{touvron2023llama}
Hugo Touvron, Louis Martin, Kevin Stone, Peter Albert, Amjad Almahairi, Yasmine Babaei, Nikolay Bashlykov, Soumya Batra, Prajjwal Bhargava, Shruti Bhosale, et~al.
\newblock Llama 2: Open foundation and fine-tuned chat models.
\newblock {\em arXiv preprint arXiv:2307.09288}, 2023.

\bibitem{vaswani2017attention}
Ashish Vaswani, Noam Shazeer, Niki Parmar, Jakob Uszkoreit, Llion Jones, Aidan~N Gomez, {\L}ukasz Kaiser, and Illia Polosukhin.
\newblock Attention is all you need.
\newblock {\em Advances in neural information processing systems}, 30, 2017.

\bibitem{wang2018non}
Xiaolong Wang, Ross Girshick, Abhinav Gupta, and Kaiming He.
\newblock Non-local neural networks.
\newblock In {\em Proceedings of the IEEE conference on computer vision and pattern recognition}, pages 7794--7803, 2018.

\bibitem{wu2023large}
Weijia Wu, Yuzhong Zhao, Zhuang Li, Jiahong Li, Hong Zhou, Mike~Zheng Shou, and Xiang Bai.
\newblock A large cross-modal video retrieval dataset with reading comprehension.
\newblock {\em arXiv preprint arXiv:2305.03347}, 2023.

\bibitem{xiao2021next}
Junbin Xiao, Xindi Shang, Angela Yao, and Tat-Seng Chua.
\newblock Next-qa: Next phase of question-answering to explaining temporal actions.
\newblock In {\em Proceedings of the IEEE/CVF conference on computer vision and pattern recognition}, pages 9777--9786, 2021.

\bibitem{xie2021segformer}
Enze Xie, Wenhai Wang, Zhiding Yu, Anima Anandkumar, Jose~M Alvarez, and Ping Luo.
\newblock Segformer: Simple and efficient design for semantic segmentation with transformers.
\newblock {\em Advances in neural information processing systems}, 34:12077--12090, 2021.

\bibitem{xu2017video}
Dejing Xu, Zhou Zhao, Jun Xiao, Fei Wu, Hanwang Zhang, Xiangnan He, and Yueting Zhuang.
\newblock Video question answering via gradually refined attention over appearance and motion.
\newblock In {\em Proceedings of the 25th ACM international conference on Multimedia}, pages 1645--1653, 2017.

\bibitem{xu2016msr}
Jun Xu, Tao Mei, Ting Yao, and Yong Rui.
\newblock Msr-vtt: A large video description dataset for bridging video and language.
\newblock In {\em Proceedings of the IEEE conference on computer vision and pattern recognition}, pages 5288--5296, 2016.

\bibitem{xu2024pllava}
Lin Xu, Yilin Zhao, Daquan Zhou, Zhijie Lin, See~Kiong Ng, and Jiashi Feng.
\newblock Pllava: Parameter-free llava extension from images to videos for video dense captioning.
\newblock {\em arXiv preprint arXiv:2404.16994}, 2024.

\bibitem{yang2022zero}
Antoine Yang, Antoine Miech, Josef Sivic, Ivan Laptev, and Cordelia Schmid.
\newblock Zero-shot video question answering via frozen bidirectional language models.
\newblock {\em Advances in Neural Information Processing Systems}, 35:124--141, 2022.

\bibitem{yi2019clevrer}
Kexin Yi, Chuang Gan, Yunzhu Li, Pushmeet Kohli, Jiajun Wu, Antonio Torralba, and Joshua~B Tenenbaum.
\newblock Clevrer: Collision events for video representation and reasoning.
\newblock {\em arXiv preprint arXiv:1910.01442}, 2019.

\bibitem{zhang2023video}
Hang Zhang, Xin Li, and Lidong Bing.
\newblock Video-llama: An instruction-tuned audio-visual language model for video understanding.
\newblock {\em arXiv preprint arXiv:2306.02858}, 2023.

\bibitem{zhang2023llama}
Renrui Zhang, Jiaming Han, Chris Liu, Peng Gao, Aojun Zhou, Xiangfei Hu, Shilin Yan, Pan Lu, Hongsheng Li, and Yu Qiao.
\newblock Llama-adapter: Efficient fine-tuning of language models with zero-init attention.
\newblock {\em arXiv preprint arXiv:2303.16199}, 2023.

\bibitem{zhang2024llavanextvideo}
Yuanhan Zhang, Bo Li, haotian Liu, Yong~jae Lee, Liangke Gui, Di Fu, Jiashi Feng, Ziwei Liu, and Chunyuan Li.
\newblock Llava-next: A strong zero-shot video understanding model, April 2024.

\bibitem{zheng2024judging}
Lianmin Zheng, Wei-Lin Chiang, Ying Sheng, Siyuan Zhuang, Zhanghao Wu, Yonghao Zhuang, Zi Lin, Zhuohan Li, Dacheng Li, Eric Xing, et~al.
\newblock Judging llm-as-a-judge with mt-bench and chatbot arena.
\newblock {\em Advances in Neural Information Processing Systems}, 36, 2024.

\bibitem{zhou2018towards}
Luowei Zhou, Chenliang Xu, and Jason Corso.
\newblock Towards automatic learning of procedures from web instructional videos.
\newblock In {\em Proceedings of the AAAI Conference on Artificial Intelligence}, volume~32, 2018.

\bibitem{zhu2023minigpt}
Deyao Zhu, Jun Chen, Xiaoqian Shen, Xiang Li, and Mohamed Elhoseiny.
\newblock Minigpt-4: Enhancing vision-language understanding with advanced large language models.
\newblock {\em arXiv preprint arXiv:2304.10592}, 2023.

\end{thebibliography}
}

\end{document}